%% file: iclr2024_conference.tex
\documentclass{article} 
\usepackage{iclr2024_conference,times}

\input{math_commands.tex}

\usepackage{hyperref}
\usepackage{url}
\iclrfinalcopy

\usepackage{booktabs}

\usepackage{multirow}
\usepackage{adjustbox}
\usepackage{amsmath}
\usepackage{color}

\usepackage{graphicx}
\usepackage{caption}
\usepackage{subcaption}
\usepackage{listings}
\usepackage{enumerate}
\usepackage{enumitem}
\usepackage{makecell}

\usepackage{color}
\usepackage{tikz}
\usepackage{pgf}
\usepackage{pgfplots}
\usetikzlibrary{patterns}
\usepackage{tikz-qtree}
\usetikzlibrary{arrows,decorations.pathmorphing,backgrounds,positioning,fit,petri,shapes.misc, arrows.meta,shapes.geometric,decorations.markings,calc,shadows.blur,decorations.pathreplacing,quotes,matrix,shapes.symbols}
\usepgfplotslibrary{colormaps}
\usepackage{epstopdf}

\usetikzlibrary{arrows,decorations.pathmorphing,backgrounds,positioning,fit,petri,shapes.misc, arrows.meta,shapes.geometric,decorations.markings,calc,shadows.blur,decorations.pathreplacing,quotes,matrix,shapes.symbols}

\definecolor{fontgray}{RGB}{44, 62, 80}
\definecolor{myred}{RGB}{235, 47, 6} 
\definecolor{summertime}{RGB}{245, 205, 121}
\definecolor{darkgrass}{RGB}{0, 148, 50}
\definecolor{myblue}{RGB}{0, 168, 255}
\definecolor{mygray}{RGB}{158, 158, 158}
\definecolor{puffin}{RGB}{250, 152, 58}
\definecolor{lowpurple}{RGB}{210, 180, 222}
\definecolor{lowblue}{RGB}{102,178,255}
\definecolor{lowred}{RGB}{245, 183, 177}
\definecolor{deeppurple}{RGB}{142, 68, 173}
\definecolor{nephritis}{RGB}{39, 174, 96}

\definecolor{deepblue}{RGB}{41, 128, 185}
\definecolor{shymoment}{RGB}{162, 155, 254}
\definecolor{firstdate}{RGB}{250, 177, 160}
\definecolor{mintleaf}{RGB}{0, 184, 148}
\definecolor{alizarin}{RGB}{231, 76, 60}
\definecolor{soaring}{RGB}{149, 175, 192}
\definecolor{electronblue}{RGB}{9, 132, 227}
\definecolor{pinkgla}{RGB}{0, 184, 148}
\definecolor{coral}{RGB}{255, 127, 80}

\newcommand{\squishlist}{
	\begin{list}{$\bullet$}
		{ \setlength{\itemsep}{0pt}
			\setlength{\parsep}{3pt}
			\setlength{\topsep}{3pt}
			\setlength{\partopsep}{0pt}
			\setlength{\leftmargin}{1.5em}
			\setlength{\labelwidth}{1em}
			\setlength{\labelsep}{0.5em} } }

	\newcounter{Lcount}
	\newcommand{\squishlisttwo}{
		\begin{list}{\arabic{Lcount}. }
			{ \usecounter{Lcount}
				\setlength{\itemsep}{0pt}
				\setlength{\parsep}{0pt}
				\setlength{\topsep}{0pt}
				\setlength{\partopsep}{0pt}
				\setlength{\leftmargin}{2em}
				\setlength{\labelwidth}{1.5em}
				\setlength{\labelsep}{0.5em} } }
		
		\newcommand{\squishend}{
	\end{list} }

\hypersetup{
  colorlinks,
  linkcolor={red!50!black},
  citecolor={blue!50!black},
  urlcolor={blue!80!black}
  }

\title{Design of Chain-of-Thought in Math Problem Solving}

\author{Zhanming Jie$^{*}$, Trung Quoc Luong$^{*}$, Xinbo Zhang\thanks{Equal contribution, order decided by the initial letter of the surname.}~~, Xiaoran Jin, Hang Li \\
ByteDance Research\\
\texttt{\{allan, trung.luong, zhangxinbo.freya\}@bytedance.com} \\
\texttt{\{xiaoran.jin, lihang.lh\}@bytedance.com} 
}

\pgfplotsset{compat=1.18} 
\begin{document}

\maketitle

\begin{abstract}

Chain-of-Thought (CoT) plays a crucial role in reasoning for math problem solving. We conduct a comprehensive examination of methods for designing CoT, comparing conventional natural language CoT with various program CoTs, including the \textit{self-describing program}, the \textit{comment-describing program}, and the \textit{non-describing program}.
Furthermore, we investigate the impact of programming language on program CoTs, comparing \textit{Python} and \textit{Wolfram Language}.
Through extensive experiments on \textsc{GSM8K}, \textsc{MathQA}, and \textsc{SVAMP}, we find that program CoTs often have superior effectiveness in math problem solving.
Notably, the best performing combination with $30$B parameters beats {GPT-3.5-turbo} by a significant margin. The results show that self-describing program offers greater diversity and thus can generally achieve higher performance. 
We also find that Python is a better choice of language than Wolfram for program CoTs.
The experimental results provide a valuable guideline for future CoT designs that take into account both programming language and coding style for further advancements. 
Our datasets and code are publicly available\footnote{\url{https://github.com/lqtrung1998/mwp_cot_design}}.

\end{abstract}

\section{Introduction}
\label{sec:intro}
\input{010intro}

\section{Chain-of-Thought Design}
\label{sec:representation}
\input{020representation}

\section{Data Collection}
\label{sec:data}
\input{030data}

\section{Methodology}
\label{sec:model}

\input{040model}

\section{Experiments}
\label{sec:exp}

\input{050experiments}

\section{Analysis}
\label{sec:analysis}
\input{051analysis}

\section{Related Work}
\label{sec:related}
\input{060related}

\section{Conclusion}
\label{sec:conclusion}
\input{070conclusion}

\bibliography{iclr2024_conference}
\bibliographystyle{iclr2024_conference}

\newpage
\appendix
\section{Hyperparameters}
\input{100appendix}

\section{Dataset Processing}
\label{sec:appendix_data}
We preprocess the original MathQA~\citep{amini2019mathqa} dataset\footnote{\url{https://huggingface.co/datasets/math_qa}} to filter out some invalid instances that contains incorrect answers~\citep{tan2022investigating,jie2022learning}. 
For example, some of the annotated equations do not lead to the correct answer in MathQA.
For SVAMP, the training set comes from the original implementation~\citep{patel2021nlp}\footnote{\url{https://github.com/arkilpatel/SVAMP}}.

\section{RM-Weighted Voting}
\label{sec:rmweighted}

\begin{table}[ht!]
    \centering
    \adjustbox{max width=1.0\linewidth}{
    \begin{tabular}{clrccc}
         \toprule
           \multicolumn{1}{c}{\bf Program} & \textbf{Method} &\multirow{1}{*}{\bf Size}	& \multicolumn{1}{c}{\bf GSM8K} & \multicolumn{1}{c}{\bf MathQA} & \multicolumn{1}{c}{\bf SVAMP} \\
           \midrule
          \multirow{1}{*}{-} & {SFT} + RM-Weighted Voting + Natural Language & $6.7$B &  $58.0$ & $64.0$  &$67.7$  \\
          \midrule
          
          \multirow{3}{*}{{Wolfram}} & {SFT} + RM-Weighted Voting + Non-Describing Program & $6.7$B &$59.4$  & $71.0$& $65.9$ \\
          & {SFT} + RM-Weighted Voting + Self-Describing Program & $6.7$B & $68.2$ & $73.1$  & $75.9$ \\
          & {SFT} + RM-Weighted Voting + Comment-Describing Program & $6.7$B &  $68.8$ & $71.6$  & $73.7$ \\
          \midrule
          
          \multirow{3}{*}{{Python}} & {SFT} + RM-Weighted Voting + Non-Describing Program & $6.7$B & $63.1$ & $70.7$ & $68.8$ \\
          & {SFT} + RM-Weighted Voting + Self-Describing Program & $6.7$B & $70.6$ & $74.8$ & $78.2$\\
          & {SFT} + RM-Weighted Voting + Comment-Describing Program & $6.7$B & $67.6$  & $71.5$ & $69.3$\\
          \midrule
          \multirow{1}{*}{-} & {SFT} + RM-Weighted Voting + Natural Language & $30$B & $71.2$& $72.8$ & $77.0$\\
          \midrule
          
          \multirow{3}{*}{{Wolfram}} & {SFT} + RM-Weighted Voting + Non-Describing Program & $30$B & $71.4$ & $72.9$ & $76.8$\\
          & {SFT} + RM-Weighted Voting + Self-Describing Program & $30$B & $76.8$ & $78.3$ & $82.5$ \\
          & {SFT} + RM-Weighted Voting + Comment-Describing Program & $30$B & $75.9$ & $74.0$ & $81.4$ \\
          \midrule
          
          \multirow{3}{*}{{Python}}& {SFT} + RM-Weighted Voting + Non-Describing Program & $30$B & $74.4$ &$73.0$  & $78.4$ \\
           & {SFT} + RM-Weighted Voting + Self-Describing Program & $30$B &  $79.5$& $77.9$& $86.0$\\
          & {SFT} + RM-Weighted Voting + Comment-Describing Program & $30$B &$77.8$ & $75.1$ & $81.2$\\
        \bottomrule
    \end{tabular}
    }
    \caption{Performance of majority voting weighted by reward model.}
    \label{tab:rm_weighted_reranking}
\end{table}

\end{document}

%% file: math_commands.tex
\usepackage{amsmath,amsfonts,bm}

\def\eqref#1{equation~\ref{#1}}

\def\1{\bm{1}}

\DeclareMathAlphabet{\mathsfit}{\encodingdefault}{\sfdefault}{m}{sl}
\SetMathAlphabet{\mathsfit}{bold}{\encodingdefault}{\sfdefault}{bx}{n}



%% file: 010intro.tex
\begin{figure}[t!]
  \centering
        \centering
        \makebox[\textwidth][c]{\includegraphics[width=\textwidth]{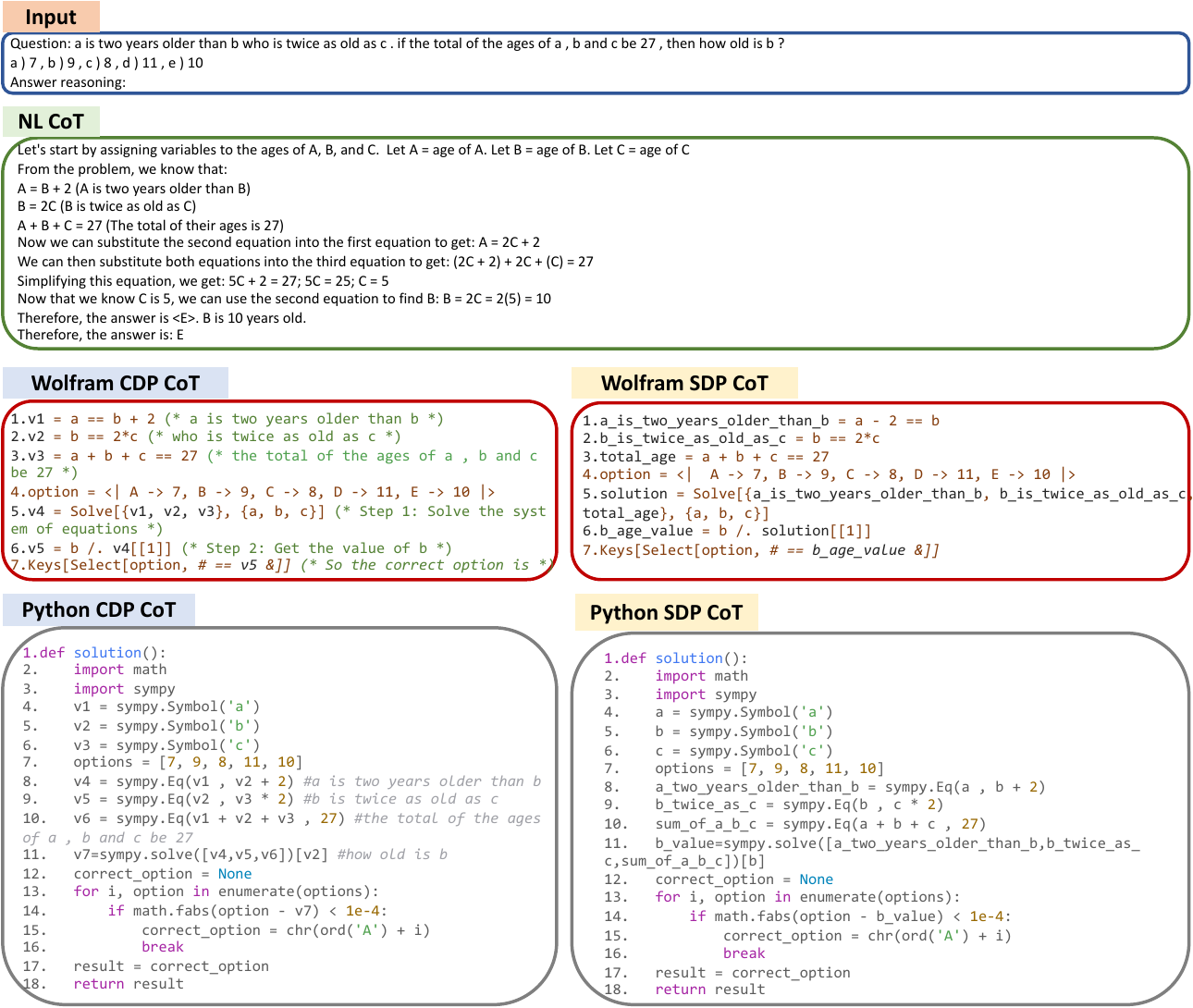}}
 \vspace*{-1mm}
  \caption{Examples of CoT representations: Natural Language (\textsc{NL}) CoT, Comment-Describing Program (\textsc{CDP}) and Self-Describing Program (\textsc{SDP}) in both Wolfram and Python.}
  \label{cdp_pal_nl_cot}
\end{figure}

Math problem solving is an ideal task to assess the multi-step reasoning abilities of large language models (LLMs).
LLMs exhibit remarkable reasoning abilities with the use of chains-of-thought, surpassing previous methods in various reasoning tasks~\citep{lightman2023lets,wei2022chain,touvron2023llama}.
The challenge of producing reliable chains-of-thought (CoT)~\citep{wei2022chain} remains, however, particularly in the nuanced and complex cases of mathematical problem solving~\citep{golovneva2022roscoe}.
Recent research has focused on refining prompt engineering strategies or developing new CoT representations, such as program CoTs~\citep{gao2023pal,he2023solving}.
Although existing approaches can boost overall performance~\citep{lu2022survey}, a thorough comparison of various CoTs remains absent in the literature.
In this paper, we conduct a comprehensive examination of multiple CoT designs, including natural language (\textsc{NL}) and program CoTs, such as the \textit{self-describing program}, the \textit{comment-describing program}, and the \textit{non-describing program}.
Figure \ref{cdp_pal_nl_cot} illustrates different CoT representations for solving a multi-choice math problem.
For program CoTs, besides the popular programming language \textit{Python}, we also use \textit{Wolfram Language}~\citep{wolframlang},  a scientific programming language known for its ability to naturally express complex mathematical expressions.

One advantage of the program CoTs is that their validity can be easily verified by executing the programs. For instance, we can easily represent an equation in one line (e.g., Figure \ref{cdp_pal_nl_cot}  Wolfram line $1$) and solve it with built-in functions (i.e., \texttt{Solve[]}). 
The \textsc{NL} CoTs do not have this capability, but they can better translate the questions in language into descriptions of reasoning by leveraging the power of LLMs. We consider three types of program CoTs.
The self-describing program (\textsc{SDP}) is similar to PAL~\citep{gao2023pal} in which the variable names are extracted from the questions.
In contrast, the non-describing program (\textsc{NDP}) only uses abstract variable names (e.g.,  $v1$ and $v2$). 
In \textsc{SDP}, programs can be created more easily from the questions, while in \textsc{NDP}, programs can be used more effectively in reasoning.
To combine the strengths of both types, we introduce the comment-describing program (\textsc{CDP}), a new CoT design that blends abstract variable names with natural language comments.

Following the common practice~\citep{uesato2022solving,lightman2023lets}, we conduct fine-tuning, reranking, and majority-voting experiments to compare the CoTs on \textsc{GSM8K}~\citep{cobbe2021training}, \textsc{MathQA}~\citep{amini2019mathqa}, and \textsc{SVAMP}~\citep{patel2021nlp} datasets.
Under the best setting, the method using the 30B model with reward model reranking is able to outperform the GPT-3.5-turbo's few-shot performance by approximately $2.9$\% on \textsc{GSM8K}, $18$\% on \textsc{MathQA} and $8$\% on \textsc{SVAMP}. We make the following main conclusions from the experiments.

\squishlisttwo
\item Program CoTs generally perform better than natural language CoTs, indicating that the use of more rigid CoTs is better.
\item The presence of natural language in SDP and CDP is crucial for achieving high performance compared with NDP. SDP is generally superior to CDP, because it can generate more diverse CoTs and thus achieve higher performance in majority voting and reranking.
\item Program CoTs in Python perform better than those in Wolfram, when the CoTs are in the same type.
\item By combining the use of different types of CoTs, we can enhance overall performance, showing the potential for further CoT design that takes advantage of the strengths of all CoT types.
\squishend
Our findings offer valuable insights for designing CoTs in math problem solving and more broadly reasoning with LLMs.

%% file: 020representation.tex
\label{section:representation}

\subsection{Natural Language CoTs (NL)}
\cite{wei2022chain} propose the chain-of-thought prompting technique to enhance the complex reasoning abilities of LLMs. 
This method endeavors to  simulate the thought process of addressing multi-step reasoning problems. 
As depicted in the second block of Figure \ref{cdp_pal_nl_cot}, this chain-of-thought approach for math problem solving produces step-by-step reasoning descriptions in natural language and provides the final answer at the end of the reasoning process. 



\subsection{Program CoTs}
We focus on two distinct programming languages: {Wolfram Language}~\citep{wolframlang} and {Python}~\citep{van1995python}.
Recent work~\citep{wang2023scibench} also uses these two languages in tool-based Transformers~\citep{schick2023toolformer}.
The Wolfram Language, with the \textit{Wolfram Mathematica} as its execution engine\footnote{\url{https://www.wolfram.com/engine/}}, is an expressive and versatile language that can effectively represent complex mathematical concepts. It has rich built-in mathematical functions for algebra, equation solving, etc., and an intuitively designed and relaxed syntax. On the other hand, Python is a general-purpose language that has gained widespread adoption in recent literature for mathematical problem solving~\citep{gao2023pal,he2023solving}. Given the contrasting nature of Wolfram and Python, we conduct a comprehensive comparison across all program CoT types in the two languages. Next, we describe the design of the CoT types, with Figure~\ref{cdp_pal_nl_cot} showcasing their instances in the two languages.



\paragraph{Self-Describing Program (SDP)}
The first design we consider is self-describing program (\textsc{SDP})  as shown in the bottom right of Figure \ref{cdp_pal_nl_cot}. 
It presents a solution in a step-by-step manner and defines variable names using natural language, similar to that of \cite{gao2023pal}. 
One advantage of \textsc{SDP} is that one can solve the problem by directly executing the program. Another advantage is that the variable names are from the question, making it easier to generate the reasoning steps for the LLM. When labeling programs, we follow several general guidelines: (1) using high-level operations to make the program concise and intuitively understandable, (2) listing variable names according to their order in the question, and (3) ensuring that variable names are meaningful, descriptive, and written in snake case naming convention (e.g., lower-cased and separated by underscores).

\paragraph{Comment-Describing Program (CDP)}
Although the design is concise, \textsc{SDP} has several problems. The self-describing names may not be sufficiently general across problems and sufficiently informative to provide rich context in CoTs. 
Therefore, we consider comment-describing program (\textsc{CDP}) using standardized variable names, e.g., $v_1$, $v_2$, and brief comments that describe the step of reasoning and problem solving.
Figure \ref{cdp_pal_nl_cot} (bottom left) shows an example in Python and Wolfram. The comment in a declaration line is a brief problem statement that provides details. The comment in a reasoning line explains the purpose of the step, displayed as a command or instruction. 
Since the Python language often requires stricter syntax, extra declaration lines, such as the Sympy symbol declaration line, must be included in the program to make it executable. 
In such lines, the comment is omitted. 

\paragraph{Non-Describing Program (NDP)}
We also consider a variant where the comments of \textsc{CDP} are discarded. \textsc{NDP} can also be considered as an approach contrary to \textsc{SDP} whereas in the former variable names are defined in natural language and in the latter variable names are defined as abstract symbols. 



%% file: 030data.tex
\begin{figure*}[t!]
  \centering
        \centering
        \makebox[\textwidth][c]{\includegraphics[width=\textwidth]{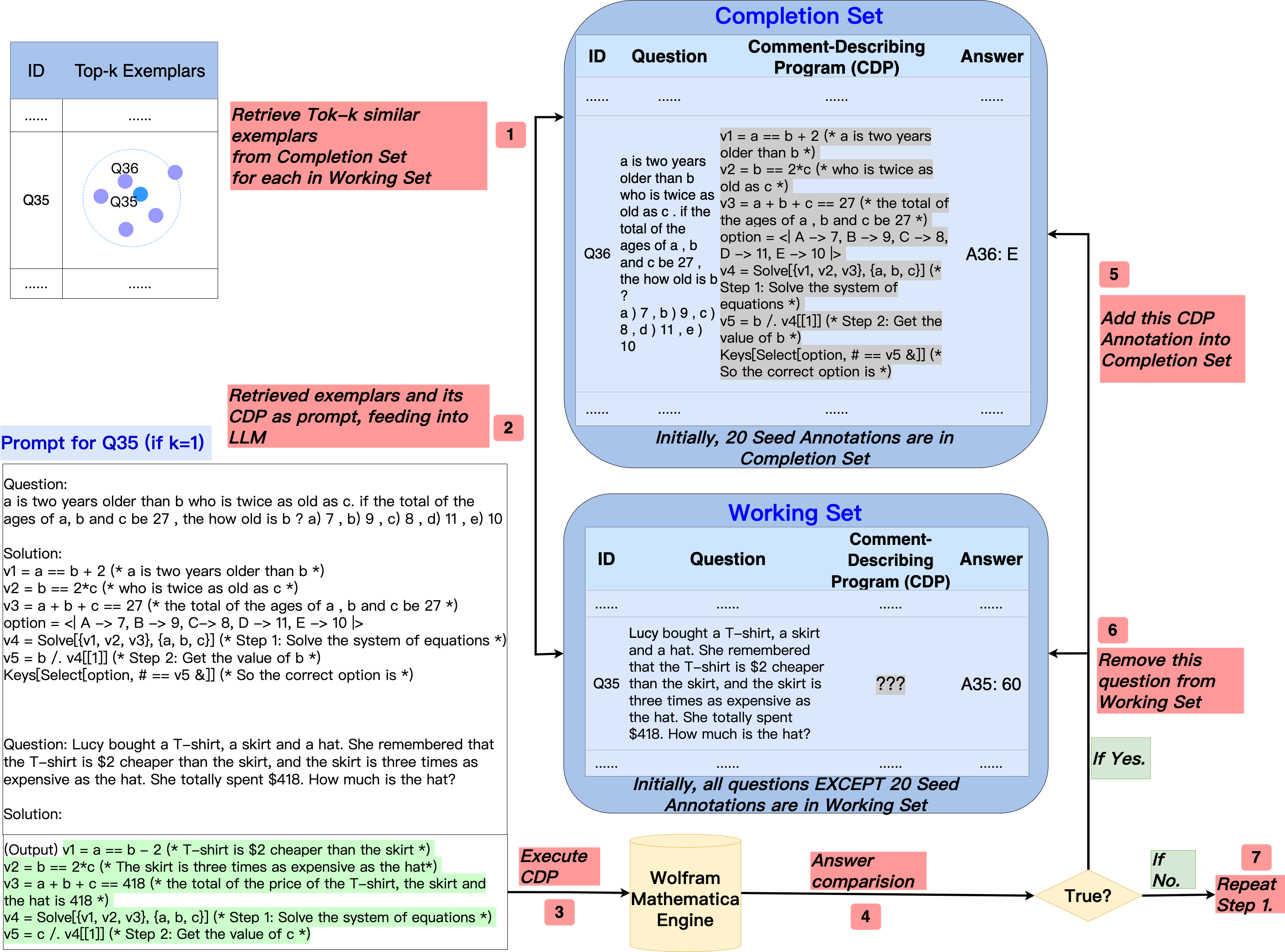}}%
 \vspace*{-1mm}
  \caption{Overview of data collection, with \textsc{CDP} as an example.}
  \label{fig:data_collection_format}
\end{figure*}

We consider three datasets in this work, \textsc{GSM8K}~\citep{cobbe2021training}, \textsc{MathQA}~\citep{amini2019mathqa}, and \textsc{SVAMP}~\citep{patel2021nlp}. 
Given the questions, we develop a method to semi-automatically annotate the CoTs in the training set.
Generally, we use the few-shot prompting technique to obtain \textsc{CDP}s and \textsc{SDP}s in both Python and Wolfram, as well as NL CoTs.  

Our LLM-empowered annotation approach works in the following way.  
We first manually create a small number of CoT annotations, and then let the LLM to retrieve similar CoT annotations as examples and generate CoTs based on the examples in a few-shot manner. 
We then automatically execute the programs and take the correctly verified annotations as the annotation results. 
The process is repeated three to five times and finally, we manually annotate those that still cannot pass the verification. 
We use the Wolfram CoT as an example to illustrate the annotation details. 

\begin{enumerate}[label=(\arabic*)]
    \item \textbf{Initial manual seed annotation} 
   We randomly select $20$ samples from the dataset for self-describing program annotations and comment-describing program annotations, respectively. The annotated programs must follow the CoT definition and Wolfram grammar. We conduct cross-verification among the authors, execute the programs by Wolfram Mathematica, and obtain the annotation results of the samples that are successfully executed. The 20 samples and their correct annotations are considered as the initial \emph{completion set}, and the other samples in the dataset are considered as the initial \emph{working set}.
    \item \textbf{Question embeddings acquisition} We acquire all the embeddings of questions in the dataset by directly calling the API of ``\texttt{text-embedding-ada-002}'' from OpenAI\footnote{\url{https://platform.openai.com/docs/guides/embeddings}}. 
\item \textbf{Retrieval-based LLM annotation}
For each sample to be annotated in the \emph{working set}, we retrieve the top-\emph{k} similar examples~\citep{liu2021makes,gao2021making} from the \emph{completion set} based on the cosine similarity of the question embeddings. 
For \textsc{CDP} annotation, we use the questions of the top-\emph{k} examples and their \textsc{CDP} programs as the prompts, and let the LLM return the \textsc{CDP} program for the given sample. 
The format of an example is presented in Fig \ref{fig:data_collection_format}. 
Here we choose ``\texttt{gpt-3.5-turbo}'' as the LLM and \emph{k} is set to $5$. 
For \textsc{SDP} annotation, we use the questions of the top-\emph{k} examples and their \textsc{SDP} programs as the prompts, and let the LLM return the \textsc{SDP} program. 
\item \textbf{Automatic verification, updating completion set and working set} After obtaining all annotations of \emph{working set} returned by the LLM, the annotated \textsc{CDP}s and \textsc{SDP}s are executed using Wolfram Mathematica, and then the results are compared with the ground truth to determine correctness. 
For \textsc{GSM8K} and \textsc{SVAMP}, since the answers should be numeric, we consider the answers not equal unless they can be converted to float and their values differ at most by $1e^{-3}$. 
For \textsc{MathQA}, due to the multiple-choice format of questions, we adopt exact match to compare answers. 
Samples with correct results after execution are put into the \emph{completion set}, and are removed from the \emph{working set}.
\item Repeat step $3$ and step $4$ for three to five times until the \emph{working set} is empty or no new samples can be added into the \emph{completion set}.
\item \textbf{Manually modifying remaining working set}
If there are still any remaining samples in the \emph{working set}, we manually conduct annotations on the samples, until the programs can get correct results using Wolfram Mathematica.
\end{enumerate}
The ways of creating Python CoTs and \textsc{NL} CoTs are the same as above.
Note that for \textsc{NL} CoTs, because they cannot be directly verified by an engine, we just apply a simple rule followed by \textsc{NL} CoT, \textit{``Therefore the answer is:"}, to obtain the answers.
\textsc{NDP}s in Wolfram and Python can be obtained by removing the comments in their corresponding \textsc{CDP}s.

%% file: 040model.tex

In accordance with previous studies \citep{uesato2022solving, lightman2023lets}, we employ supervised fine-tuning (SFT), \textit{self-consistency} decoding (alternatively referred to as \textit{majority voting}) \citep{wang2022self}, and reward model \textit{reranking} methodologies on our annotated dataset.

\subsection{Supervised Fine-tuning}
We conduct SFT on a pre-trained language model using questions and chain-of-thought annotations in each dataset. 
The training aims to maximize the likelihood of the answer given the question. 
In evaluation, we extract the final answer generated by the SFT model.
As shown in Figure \ref{cdp_pal_nl_cot}, the \textsc{NL} CoT places the final answer in the last sentence, ``\textit{Therefore, the answer is E.}''.
In the cases of \textsc{SDP}, \textsc{CDP}, and \textsc{NDP}, we execute the program to obtain the answer.

\subsection{Majority Voting}
In self-consistency decoding~\citep{wang2022self}\footnote{We use the term ``majority voting'' in the rest of this paper unless specified.}, we first sample a certain number of CoTs from the language model.
We then perform majority voting over the answers extracted from all the sampled CoTs and choose the final answer that is the most favored among all answers. We simply adopt the temperature sampling strategy~\citep{ackley1985learning,ficler2017controlling} with $T= 1.0$, because it is reported~\citep{wang2022self} that self-consistency decoding is generally robust to sampling strategies and hyperparameters.

\subsection{Reranking with Reward Model}
\label{sec:reranking}
Following \cite{cobbe2021training}, a reward model (RM) is trained to determine whether an answer to the question is correct or not. 
Given the SFT model, we perform sampling to obtain a certain number of CoT solutions to the question. As a common practice, the reward model is a language model that is initialized from the SFT model. 
Similar to the outcome-based reward model (ORM)~\citep{uesato2022solving}, the reward model is trained to predict a binary label that indicates the ``\textit{correct}'' or ``\textit{incorrect}'' solution\footnote{Our preliminary experiments with process-based reward~\citep{lightman2023lets} show similar performance with outcome-based reward.
We attribute the reason to the quality of automatic process labels~\citep{uesato2022solving}. 
}.
Once the input passes through the reward model, classification is conducted with a linear classifier on the hidden state of the last token. 
Finally, the solution with the highest ``\textit{correct}'' score among the candidates is selected as the final answer.

As we do not have explicit ``\textit{correct}'' and ``\textit{incorrect}'' pairs annotated, we adopt the model at the 2$^{nd}$ epoch during supervised fine-tuning to sample solution pairs.
According to \cite{cobbe2021training}, using the checkpoints at initial epochs can provide more diverse solutions for training a reward model.
For each question in the training data $\mathcal{D}$, we sample $K$ solutions. We then use all the samples that contain both correct and incorrect solutions to train the reward model for three epochs.

%% file: 050experiments.tex
\subsection{Experiment Settings}
We conduct experiments on the three datasets: \textsc{GSM8K}~\citep{cobbe2021training}, \textsc{MathQA}~\citep{amini2019mathqa}\footnote{Due to limitation of computing budget, we only experiment with a random $15k$ samples of \textsc{MathQA} training set.}, and \textsc{SVAMP}~\citep{patel2021nlp}. 
Appendix \S\ref{sec:appendix_data} illustrates the preprocessing procedure for MathQA and SVAMP dataset. 
The training data for all CoT types is obtained using the method described in \S\ref{sec:data}.
We report the results of few-shot prompting using {GPT-3.5-turbo}, majority voting, and RM reranking\footnote{We also experimented with RM-weighted voting and the performance is similar to reranking (Appendix \S\ref{sec:rmweighted}).}. 

We adopt the pre-trained language model Galactica~\citep{taylor2022galactica} which is trained on a large-scale scientific corpus and programming codes. 
The Galactica model shows superior performance in math problem solving compared to other foundation models such as LLaMA~\citep{touvron2023llama} in our preliminary experiments. 
Throughout the experiments, we use the model size of $6.7$B\footnote{\url{https://huggingface.co/facebook/galactica-6.7b}} and $30$B\footnote{\url{https://huggingface.co/facebook/galactica-30b}} available in HuggingFace. 
We use the Megatron-Deepspeed\footnote{\url{https://github.com/bigscience-workshop/Megatron-DeepSpeed}} framework for efficient supervised fine-tuning, following BLOOM~\citep{scao2022bloom}.
The model is fine-tuned for $40$ epochs with a maximum sequence length of $1024$. Please refer to Appendix (Table \ref{tab:SFT and RM hyperparameter}) for hyper-parameter settings.

We select the SFT model with the best accuracy for sampling to obtain the majority voting~\citep{wang2022self} results.
To train the reward model, we generate $100$ samples for each question in the training set using the SFT  checkpoint at the second epoch and compare them with the ground-truths to determine the correct labels.  
By using an earlier checkpoint, we can have more sampling diversity, which is helpful for the reward model training.
As described in \S\ref{sec:reranking}, we initialize the reward model with the best SFT model checkpoint and fine-tune it for three epochs.

\begin{table}[t!]
    \centering
    \adjustbox{max width=0.85\linewidth}{
    \begin{tabular}{clrcccc}
         \toprule
          \multirow{1}{*}{\bf Program}&\multirow{1}{*}{\textbf{CoT Type}} &\multirow{1}{*}{\textbf{Size}} & \multicolumn{1}{c}{\bf GSM8K} & \multicolumn{1}{c}{\bf MathQA} & \multicolumn{1}{c}{\bf SVAMP}& \multicolumn{1}{c}{\bf Avg.} \\
          \midrule
          - & Natural Language &  $6.7$B & $41.0$ & $58.7$ & $53.8$ & $51.2$\\
          \midrule
          \multirow{3}{*}{Python} & Non-Describing Program &\multirow{1}{*}{$6.7$B}& $56.3$ & $64.4$ & $59.1$ & $59.9$\\
          &   Self-Describing Program & $6.7$B & $\textbf{57.1}$ & $\textbf{64.8}$ & $\textbf{69.3}$ & $\textbf{63.7}$ \\
          & Comment-Describing Program &  $6.7$B  & $56.5$ & $64.7$ & $62.3$ & $61.2$ \\
          \midrule
          \multirow{2}{*}{Wolfram}& Non-Describing Program &\multirow{1}{*}{$6.7$B}& $53.4$ & $63.0$& $58.6$ & $58.3$ \\
           & Self-Describing Program &  $6.7$B  & $50.2$ & $62.5$ & $\textbf{65.5}$ & $59.4$ \\
          & Comment-Describing Program & $6.7$B   & $\textbf{57.0}$ & $\textbf{63.1}$ & $64.0$ & $\textbf{61.4}$ \\
          \midrule
          \midrule
          - & Natural Language & $30$B & $57.4$ & $66.6$ & $70.1$ & $64.7$\\
          \midrule
          \multirow{3}{*}{Python} 
          & Non-Describing Program &\multirow{1}{*}{$30$B}& $65.8$ & $66.0$ & $73.9$ & $68.6$\\
          & Self-Describing Program & $30$B &  $68.3$ & $67.2$ & $\textbf{80.4}$ & $\textbf{72.0}$\\
          & Comment-Describing Program & $30$B&  $\textbf{68.7}$ & $\textbf{67.2}$ & $78.2$ & $71.4$ \\
          \midrule
          \multirow{3}{*}{Wolfram} 
          & Non-Describing Program &\multirow{1}{*}{$30$B}& $62.2$ & $64.9$ & $73.1$ & $66.7$\\
          & Self-Describing Program & $30$B & $62.6$ & $64.3$ & $73.9$ & $66.9$ \\
          & Comment-Describing Program & $30$B & $\textbf{66.7}$ & $\textbf{65.0}$ & $\textbf{75.9}$ & $\textbf{69.2}$ \\
        \bottomrule
    \end{tabular}
    }
    \caption{Supervised fine-tuning performance of all CoT types. Numbers displayed in bold are highest in the same settings.}
    \label{tab:sft_results}
\end{table}

\subsection{Supervised Fine-tuning Results}
Table \ref{tab:sft_results} presents the supervised fine-tuning results across all datasets, languages, and CoT types.
In general, program-based CoTs perform better than natural language CoT.
An enlarged model size correlates with a noticeable increase in the performance of natural language CoTs, presumably due to an improved capacity for natural language understanding.
Nevertheless, program-based CoTs consistently and significantly outperform natural language CoT.
The SFT models described here are then used for majority voting and reranking.

\begin{table}[ht!]
    \centering
    \adjustbox{max width=1.0\linewidth}{
    \begin{tabular}{clrccc}
         \toprule
           \multicolumn{1}{c}{\bf Program} & \textbf{Method} &\multirow{1}{*}{\bf Size}	& \multicolumn{1}{c}{\bf GSM8K} & \multicolumn{1}{c}{\bf MathQA} & \multicolumn{1}{c}{\bf SVAMP} \\
           \midrule
           \midrule
          - & {GPT-3.5-turbo}$_{~\text{\footnotesize prompting}}$ + Natural Language & N.A. & $75.3$ & $60.6$ & $73.0$ \\
          \midrule
          \multirow{2}{*}{{Wolfram}} & {GPT-3.5-turbo}$_{~\text{\footnotesize prompting}}$ + Self-Describing Program & N.A. & $73.5$ & $39.3$ & $72.8$\\
          & {GPT-3.5-turbo}$_{~\text{\footnotesize prompting}}$ + Comment-Describing Program & N.A. & $69.1$ & $31.2$ & $70.1$\\
          \midrule
          \multirow{2}{*}{{Python}} & {GPT-3.5-turbo}$_{~\text{\footnotesize prompting}}$ + Self-Describing Program & N.A. & $78.0$ & $45.5$ & $78.4$\\
          & {GPT-3.5-turbo}$_{~\text{\footnotesize prompting}}$ + Comment-Describing Program & N.A. & $72.4$ & $46.2$ & $77.6$\\
          \midrule
          \midrule
          \multirow{2}{*}{-} & {SFT} + Majority Voting + Natural Language & $6.7$B & $50.8$ & $ 59.5$ & $63.1$ \\
          & SFT + Reranking + Natural Language & $6.7$B & $59.5$ & $61.8$ & $67.0$\\
          \midrule
          
          \multirow{6}{*}{{Wolfram}} & {SFT} + Majority Voting + Non-Describing Program & $6.7$B & $53.7$ & $70.3$ & $60.9$ \\
          & {SFT} + Majority Voting + Self-Describing Program & $6.7$B & $59.5$ & $72.7$ & $72.9$\\
          & {SFT} + Majority Voting + Comment-Describing Program & $6.7$B & $61.3$ & $71.3$ & $68.3$ \\
          & SFT + Reranking + Non-Describing Program & $6.7$B & $61.1$ & $71.0$ & $66.4$\\
          & SFT + Reranking + Self-Describing Program & $6.7$B & $71.4$& $73.8$ & $77.3$\\
          & SFT + Reranking + Comment-Describing Program & $6.7$B & $69.7$ & $72.0$ & $75.5$\\
          \midrule
          
          \multirow{6}{*}{{Python}} & {SFT} + Majority Voting + Non-Describing Program & $6.7$B & $56.4$ & $70.5$ & $61.6$ \\
          & {SFT} + Majority Voting + Self-Describing Program & $6.7$B & $61.1$ & $73.7$ & $73.7$\\
          & {SFT} + Majority Voting + Comment-Describing Program & $6.7$B & $58.6$ & $71.5$ & $63.4$ \\
          & SFT + Reranking + Non-Describing Program & $6.7$B & $63.6$ & $70.7$& $69.7$\\
          & SFT + Reranking + Self-Describing Program & $6.7$B & $\mathbf{72.4}$ & $\mathbf{75.2}$ & $\mathbf{78.6}$\\
          & SFT + Reranking + Comment-Describing Program & $6.7$B & $69.9$ & $71.6$ & $69.8$\\
          \midrule
          \midrule
          \multirow{2}{*}{-} & {SFT} + Majority Voting + Natural Language & $30$B & $69.8$ &$69.4$ & $72.0$\\
          & SFT + Reranking + Natural Language & $30$B & $75.7$ & $74.3$ & $77.0$\\
          \midrule
          
          \multirow{6}{*}{{Wolfram}} & {SFT} + Majority Voting + Non-Describing Program & $30$B & $63.4$ & $72.2$ & $73.3$ \\
          & {SFT} + Majority Voting + Self-Describing Program & $30$B & $69.8$ & $77.9$ & $78.7$\\
          & {SFT} + Majority Voting + Comment-Describing Program & $30$B & $69.8$ & $73.6$ & $79.2$\\
          & SFT + Reranking + Non-Describing Program & $30$B & $71.4$ & $72.9$ & $77.2$\\
          & SFT + Reranking + Self-Describing Program & $30$B & $79.9$ & $\mathbf{78.6}$ & $83.4$\\
          & SFT + Reranking + Comment-Describing Program & $30$B & $78.6$ & $74.1$ & $82.9$\\
          \midrule
          
          \multirow{6}{*}{{Python}}& {SFT} + Majority Voting + Non-Describing Program & $30$B & $67.0$ & $72.6$ & $75.3$ \\
           & {SFT} + Majority Voting + Self-Describing Program & $30$B & $72.3$ & $77.2$ & $82.7$\\
          & {SFT} + Majority Voting + Comment-Describing Program & $30$B & $70.1$ & $74.8$ & $79.1$\\
          & SFT + Reranking + Non-Describing Program & $30$B & $74.3$ & $73.0$ & $78.4$\\
          & SFT + Reranking + Self-Describing Program & $30$B & $\mathbf{80.9}$ & $78.1$ & $\mathbf{87.0}$\\
          & SFT + Reranking + Comment-Describing Program & $30$B & $78.2$ & $75.1$ & $81.5$\\
        \bottomrule
    \end{tabular}
    }
    \caption{Performance comparison among few-shot prompting, majority voting and reranking. Numbers in bold are the best and significantly better than the second-best with $p<0.001$.}
    \label{tab:reranking}
\end{table}

\subsection{Main Results}
Table \ref{tab:reranking} shows the comparison among different methods: {GPT-3.5-turbo}\footnote{\url{https://platform.openai.com/docs/models/gpt-3-5}} prompting, majority voting, and reranking. 
We can observe that larger models (i.e., $30$B) can significantly improve the performance over smaller models (i.e., $6.7$B) on all datasets. 
Our best variants are able to outperform few-shot prompting {GPT-3.5-turbo} by a large margin. 

\paragraph{Prompting Performance}
The few-shot examples are selected randomly from the training annotations for all types of CoT. Among the methods using {GPT-3.5.turbo}, natural language is generally better than comment-describing program, self-describing program, and non-describing program, while self-describing program in Python is better on \textsc{GSM8K} and \textsc{SVAMP}. 
We attribute the low performance to the limited availability of programs in the pre-training data of {GPT-3}~\citep{brown2020language}, which appears to be true for most existing LLMs~\citep{touvron2023llama,taylor2022galactica,touvron2023llama2}. 
Furthermore, it is challenging to generalize to new problems with just a few examples of in-context learning. While ongoing research addresses these challenges~\citep{min2021metaicl,hao2022structured,coda2023meta}, our work does not focus on this aspect. The CoTs of \textsc{SDP} in Python are more similar to the programming codes in the pre-training corpus , and thus the use of them leads to better performance on \textsc{GSM8K} and \textsc{SVAMP}. For the nosier dataset, \textsc{MathQA}, natural language CoTs tend to make guesses on multi-choice questions, even if they are incorrect, whereas program CoTs tend to choose no decision if there is no valid answer available.

\paragraph{Program and Natural Language Comparison}
In general, the performance with all types of program CoTs is consistently better than that with natural language CoTs. This superiority is particularly evident in the case of \textsc{MathQA} where the program CoTs, in combination with reranking, lead to performance improvement exceeding $10$ points for the $6.7$B model compared to natural language CoTs. This is because the answers are in multiple-choice format in \textsc{MathQA}, and thus inaccurate predictions for which the program execution results are ``\textit{null-result}'' can be easily filtered out before performing majority voting or reranking\footnote{We would see many predictions give ``\textit{null}'' as the voting/reranking answer if we did not remove them.}. Table \ref{tab:mathqa_percent} presents the percentage of null-result answers in \textsc{MathQA} predictions. Although natural language CoTs produce fewer null-result answers, their performance with null-result answers is worse, as shown in Table \ref{tab:mathqa_nlacc}. 
\begin{table}[ht!]
\centering
\begin{minipage}{.4\linewidth}
\centering
\begin{tabular}{lcc}
\toprule
 \multirow{2}{*}{} & \multicolumn{2}{c}{\textbf{Null-result Answer (\%)}} \\
   & $6.7$B & $30$B \\
\midrule
\textsc{NL} & \textcolor{white}{$0$}$0.53$& \textcolor{white}{$0$}$2.27$\\
\textsc{SDP} &  $34.87$&  $33.12$\\
\textsc{CDP} &  $34.73$&  $32.78$\\
\bottomrule
\end{tabular}
\caption{Percentage of \textit{null-result} answers in MathQA Wolfram predictions.}
\label{tab:mathqa_percent}
\end{minipage}
\hspace{0.1\linewidth}
\begin{minipage}{.45\linewidth}
\centering
\begin{tabular}{lcc}
\toprule
\textbf{Range} & $6.7$B&$30$B \\
\midrule
Overall & $59.5$ & $69.4$ \\
\midrule
$0\sim20$ & $73.9$ & $81.2$ \\
$20\sim40$ & $58.8$& $54.5$ \\
$40\sim60$ & $58.3$ & $62.1$\\
$60\sim80$ & $40.3$ & $55.8$ \\
$80\sim100$ & $35.7$ & $46.3$ \\
\bottomrule
\end{tabular}
\caption{NL majority voting accuracy against percentage of \textit{null-result} answers in CDP.}
\label{tab:mathqa_nlacc}
\end{minipage}
\end{table}
Therefore, it is essential to remove those samples as voting/re-ranking could be misled by them. Conversely, natural language CoTs tend to choose an answer (e.g., \texttt{A}, \texttt{B}), regardless of the accuracy of the CoT, because the CoT cannot be executed as a program. However, there are exceptions where we use the non-describing program.  For example, the performance of ``majority voting + \textsc{NDP}'' with Python using the $6.7$B model is worse than the natural language counterpart on \textsc{SVAMP}. 
The same observation also applies to the \textsc{GSM8K} dataset with the $30$B model for both Wolfram and Python languages. 
Without natural language, \textsc{NDP} has a weaker language understanding capability compared to the \textsc{SDP} and \textsc{CDP}. 

\paragraph{Program CoT Comparison}

Under both the majority voting and reranking strategies, self-describing program consistently achieve the best performance, followed by the comment-describing program, and then non-describing program. Unlike in SFT, self-describing program provides more diversity and therefore tends to perform better in voting and re-ranking. Notably, the $30$B  model with Python, ``reranking + \textsc{SDP}'', achieves the best performance on \textsc{GSM8K} and \textsc{SVAMP}. 
The performance is also $2.9$ points and $8.6$ points higher than the best prompting approach with {GPT-3.5-turbo} on \textsc{GSM8K} and \textsc{SVAMP}, respectively.``reranking + \textsc{SDP}'' with Wolfram also obtains the best performance on the noisy \textsc{MathQA} dataset, with +$28$ points improvement over {GPT-3.5-turbo} prompting. 
Though the performance with \textsc{CDP} is worse than \textsc{SDP}, we can see that the best \textsc{CDP} methods can still outperform the best {GPT-3.5-turbo} prompting approach on all datasets. 



\paragraph{Programming Language Comparison}
The best-performing $6.7$B and $30$B models are often the methods in Python, as shown in Table \ref{tab:reranking}. 
The only exception is that the best $30$B model with Python falls $0.5$ points behind the best $30$B model with Wolfram. 
For non-describing and self-describing programs, the use of Python often outperforms the use of Wolfram. 
For comment-describing program, the methods using Python and Wolfram have comparable performance, with the $6.7$B model using Wolfram having better performance on SVAMP.

%% file: 051analysis.tex



\subsection{Number of Instances for Sampling} 
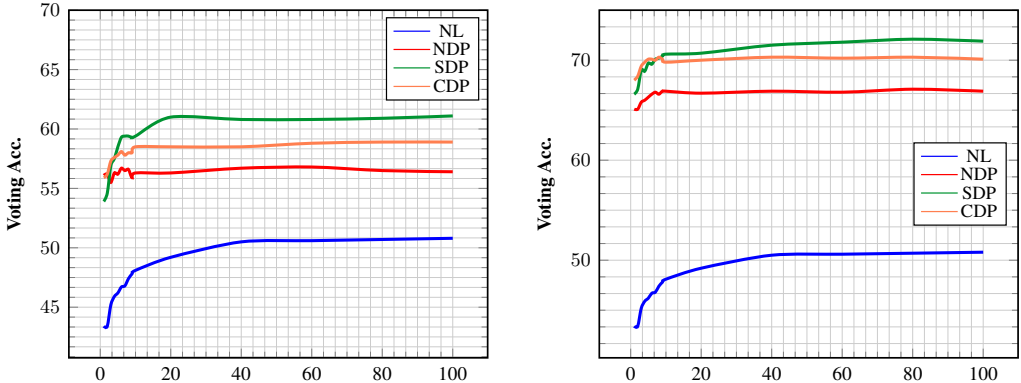
\begin{figure}[ht!]
 	\pgfplotsset{every axis/.append style={
 					ylabel=\textbf{Voting Acc.},
 			label style={font=\footnotesize},
 			tick label style={font=\footnotesize}  
 	}}
 	\begin{subfigure}[b]{0.5\textwidth}
 		\centering
 		\adjustbox{max width=0.95\textwidth}{
 			\begin{tikzpicture}
 				\begin{axis}[
 					axis line style = thick,
                         grid=both,
                         grid style={line width=.1pt, draw=gray!40},
                         minor tick num=5,
 					legend style={nodes={scale=0.5, transform shape}, line width=0.2mm}, 
 					xtick={0,20,40,60,80,100},
 					xticklabels={0,20,40,60,80,100},
 					ymax = 70, 
 					line width = 0.5mm]
 					\addplot [ smooth, solid, pattern color = blue, draw=blue] coordinates {
 						(1, 43.4)
 						(2, 43.4)
 						(3, 45.2)
 						(4, 45.9)
 						(5, 46.2)
 						(6, 46.7)
 						(7, 46.800000000000004)
 						(8, 47.4)
 						(9, 47.8)
 						(10, 48.1)
 						(20, 49.2)
 						(40, 50.5)
 						(60, 50.6)
 						(80, 50.7)
 						(100, 50.8)
 						};
 					\addlegendentry{\LARGE NL}
 					\addplot [  smooth, solid,  pattern color = red, draw=red] coordinates {
 						(1, 56.10000000000001)
 						(2, 56.2)
 						(3, 55.50000000000001)
 						(4, 56.3)
 						(5, 56.2)
 						(6, 56.699999999999996)
 						(7, 56.49999999999999)
 						(8, 56.599999999999994)
 						(9, 55.900000000000006)
 						(10, 56.3)
 						(20, 56.3)
 						(40, 56.699999999999996)
 						(60, 56.8)
 						(80, 56.49999999999999)
 						(100, 56.39999999999999)
 					};
 					\addlegendentry {\LARGE NDP};
 					\addplot [  smooth, solid, pattern color = darkgrass, draw=darkgrass] coordinates {
 						(1, 53.900000000000006)
 						(2, 54.6)
 						(3, 56.89999999999999)
 						(4, 57.49999999999999)
 						(5, 58.5)
 						(6, 59.3)
 						(7, 59.4)
 						(8, 59.4)
 						(9, 59.3)
 						(10, 59.4)
 						(20, 61.0)
 						(40, 60.8)
 						(60, 60.8)
 						(80, 60.9)
 						(100, 61.1)
 					};
 					\addlegendentry {\LARGE SDP};
 					\addplot [  smooth, solid , pattern color = coral, draw=coral] coordinates {
 						(1, 55.900000000000006)
 						(2, 56.2)
 						(3, 57.3)
 						(4, 57.599999999999994)
 						(5, 57.8)
 						(6, 58.099999999999994)
 						(7, 57.8)
 						(8, 57.99999999999999)
 						(9, 57.99999999999999)
 						(10, 58.5)
 						(20, 58.5)
 						(40, 58.5)
 						(60, 58.8)
 						(80, 58.9)
 						(100, 58.9)
 					};
 					\addlegendentry {\LARGE CDP};
 				\end{axis}
 			\end{tikzpicture}
 		}
 	\end{subfigure}
 	\begin{subfigure}[b]{0.5\textwidth}
 		\centering
 		\adjustbox{max width=0.95\textwidth}{
 			\begin{tikzpicture}
 				\begin{axis}[
 					axis line style = thick,
 					legend style={nodes={scale=0.5, transform shape}, line width=0.2mm, at={(0.75,0.5)},anchor=west}, 
                     grid=both,
                     grid style={line width=.1pt, draw=gray!40},
                     minor tick num=5,
 					xtick={0,20,40,60,80,100},
 					xticklabels={0,20,40,60,80,100},
 					ymax = 75, 
 					line width = 0.5mm]
 					\addplot [ smooth, solid, pattern color = blue, draw=blue] coordinates {
 						(1, 43.4)
 						(2, 43.4)
 						(3, 45.2)
 						(4, 45.9)
 						(5, 46.2)
 						(6, 46.7)
 						(7, 46.800000000000004)
 						(8, 47.4)
 						(9, 47.8)
 						(10, 48.1)
 						(20, 49.2)
 						(40, 50.5)
 						(60, 50.6)
 						(80, 50.7)
 						(100, 50.8)
 					};
 					\addlegendentry{\LARGE NL}
 					\addplot [  smooth, solid,  pattern color = red, draw=red] coordinates {
 						(1, 65.10000000000001)
 						(2, 65.10000000000001)
 						(3, 65.8)
 						(4, 66.0)
 						(5, 66.3)
 						(6, 66.60000000000001)
 						(7, 66.8)
 						(8, 66.60000000000001)
 						(9, 66.9)
 						(10, 66.9)
 						(20, 66.7)
 						(40, 66.9)
 						(60, 66.8)
 						(80, 67.10000000000001)
 						(100, 66.9)
 					};
 					\addlegendentry {\LARGE NDP};
 					\addplot [  smooth, solid, pattern color = darkgrass, draw=darkgrass] coordinates {
 						(1, 66.60000000000001)
 						(2, 67.10000000000001)
 						(3, 69.0)
 						(4, 68.89999999999999)
 						(5, 69.69999999999999)
 						(6, 69.6)
 						(7, 70.1)
 						(8, 70.19999999999999)
 						(9, 70.39999999999999)
 						(10, 70.6)
 						(20, 70.7)
 						(40, 71.5)
 						(60, 71.8)
 						(80, 72.1)
 						(100, 71.89999999999999)
 					};
 					\addlegendentry {\LARGE SDP};
 					\addplot [  smooth, solid , pattern color = coral, draw=coral] coordinates {
 						(1, 68.0)
 						(2, 68.4)
 						(3, 69.39999999999999)
 						(4, 69.8)
 						(5, 70.1)
 						(6, 70.1)
 						(7, 70.0)
 						(8, 70.3)
 						(9, 70.1)
 						(10, 69.8)
 						(20, 70.0)
 						(40, 70.3)
 						(60, 70.19999999999999)
 						(80, 70.3)
 						(100, 70.1)
 					};
 					\addlegendentry {\LARGE CDP};
 				\end{axis}
 			\end{tikzpicture}
 		}
 	\end{subfigure}
 	\vspace{-3mm}
 	\caption{Majority voting regarding the different number of sampled instances (Left: $6.7$B; Right: $30$B). We just depict the performance in Python for illustration purposes.}
 	\label{fig:analysis_k}
 \end{figure}

We measure the effect of the number of sampled instances $K$ during majority voting. We vary the number $K$ from $1$ to $100$ and evaluate the accuracies for the $6.7$B and $30$B models. Figure \ref{fig:analysis_k} provides the results on \textsc{GSM8K}. 
The performance of all methods improves rapidly with an increase of $K$ and becomes stable when $K$ is more than $30$. 
Specifically, both \textsc{CDP} and \textsc{NDP} require a smaller number of $K$ compared to \textsc{SDP} and \textsc{NL}. 
The results indicate that \textsc{CDP} and \textsc{NDP} are more deterministic while \textsc{SDP} and \textsc{NL} are more diverse. 
With more diverse CoTs, \textsc{SDP} and \textsc{NL} are able to gain more improvements with more samples in majority voting. 


\subsection{Representation sampling statistics}
Table \ref{tab:sampling_statistics} reports the results of model predictions on \textsc{GSM8K}, including the percentage of syntactically correct predictions (i.e., execution rate), the percentage of correct answers (i.e., precision) and the chance of obtaining at least one correct answer among $100$ samples (i.e., correct@100). 
Here, syntactically correct means that, we can extract or execute the CoT to get a \textit{valid} answer (e.g., \texttt{A}, \texttt{B}, \texttt{C}, or \texttt{D} letter for \textsc{MathQA}, and numeric value for \textsc{GSM8K} and \textsc{SVAMP}). 

It can be seen that \textsc{NL} CoT has a high correct@100 and execution rate but with the lowest precision compared to all other CoT types.
This is probably because the natural langauge syntax is straightforward and it is challenging for the models on the current scale to perform precise calculations without the help of a computational engine.
It is noteworthy that \textsc{CDP} usually has the highest precision and execution rate, and relatively high correct@100 score, and \textsc{SDP} has the lowest execution rate but the highest correct@100 score and relatively high precision. 
The results further support our hypothesis that \textit{\textsc{CDP} is more deterministic and precise, while \textsc{SDP} has a higher level of diversity}, and thus a higher chance of obtaining correct answers with the risk of making more errors. 
Therefore, we conclude that having a balance of diversity and precision is crucial for higher performance in voting and reranking.
The execution rates of \textsc{CDP} and \textsc{NDP} are similar, but \textsc{CDP} scores higher in correct@100 and achieves significantly better precision.
Such an observation indicates the benefits of including natural language comments.
\begin{table}[ht!]
    \centering
    \adjustbox{max width=0.95\linewidth}{
    \begin{tabular}{llcccc}
         \toprule
          \textbf{Program} & \textbf{CoT Type} & \textbf{Size} & \textbf{Correct@100} & \textbf{Precision} (\%) & \textbf{Executable} (\%) \\
          \midrule
          \midrule
          - & Natural Language & $6.7$B    &$86.9$ & $41.5$ &  $99.3$  \\
          \midrule
          \multirow{3}{*}{{Wolfram}} & Comment-Describing Program & $6.7$B &$78.5$ & $58.7$ &  $99.6$  \\
           & Self-Describing Program & $6.7$B &$82.8$ & $54.7$ &  $94.1$  \\
           & Non-Describing Program & $6.7$B &$69.6$ & $52.1$ &  $99.3$  \\
           \midrule
          \multirow{3}{*}{{Python}} & Comment-Describing Program & $6.7$B &$78.8$ & $55.9$ &  $98.6$  \\
           & Self-Describing Program & $6.7$B &$83.6$ & $56.7$ &  $96.2$  \\
           & Non-Describing Program & $6.7$B &$70.9$ & $55.3$ &  $99.6$  \\
           \midrule
           \midrule
          - & Natural Language & $30$B &$94.3$ & $58.0$ &  $98.8$  \\
          \midrule
          \multirow{3}{*}{{Wolfram}} & Comment-Describing Program & $30$B  &$85.0$ & $67.7$ &  $99.6$  \\
           & Self-Describing Program & $30$B  &$91.1$ & $65.1$ &  $97.8$  \\
           & Non-Describing Program & $30$B  &$76.1$ & $62.0$ &  $99.6$  \\
           \midrule
          \multirow{3}{*}{{Python}} & Comment-Describing Program & $30$B  &$86.1$ & $68.1$ &  $99.0$  \\
          & Self-Describing Program & $30$B  &$91.0$ & $67.6$ &  $98.1$  \\
          &  Non-Describing Program & $30$B  &$82.6$ & $65.0$ &  $99.8$  \\
          \bottomrule
    \end{tabular}
    }
    \caption{Sampling statistics on \textsc{GSM8K} dataset.}
    \label{tab:sampling_statistics}
\end{table}

\subsection{Upper bounds}


We analyze the results in reranking to explore the potential of the CoT designs (\textsc{NL}, \textsc{CDP}/\textsc{SDP} in Wolfram/Python). 
We calculate the accuracy when {\it any} of the CoT is correct, which is considered the upper bound of all types of CoTs. 
We consider the best performance of individual types of CoTs and the upper bounds of all types of CoTs in Table \ref{tab:ensemble}. 
We find that the upper bounds of the CoTs on 30B models are $98.8$\%, $93.0$\%, $95.0$\% on \textsc{GSM8K}, \textsc{MathQA}, \textsc{SVAMP}, respectively. 
It indicates the potential for combining the CoTs to create a more accurate representation. We leave this as future work.

\begin{table}[ht!]
    \centering
    \adjustbox{max width=1.0\linewidth}{
    \begin{tabular}{lrccc}
         \toprule
            &\multirow{1}{*}{\bf Size}	& \multicolumn{1}{c}{\bf GSM8K} & \multicolumn{1}{c}{\bf MathQA} & \multicolumn{1}{c}{\bf SVAMP} \\
           \midrule
           \midrule
          \multirow{1}{*}{ Best performance of \textit{individual} CoT} & $6.7$B &\makecell[c]{$72.4$\\(Python SDP)}  &\makecell[c]{$75.2$\\(Python SDP)}  & \makecell[c]{$78.6$\\(Python SDP)} \\
          \midrule
          \multirow{1}{*}{ Upper bound of \textit{all} types of CoTs} & $6.7$B & $\mathbf{89.5}$ & $\mathbf{90.1}$ & $\mathbf{91.0}$ \\
          
          \midrule
           \midrule
           \multirow{1}{*}{ Best performance of \textit{individual} CoT} & $30$B &\makecell[c]{$80.9$\\(Python SDP)}  & \makecell[c]{$78.6$\\(Wolfram SDP)} & \makecell[c]{$87.0$\\(Python SDP)} \\
          
          \midrule
          \multirow{1}{*}{ Upper bound of \textit{all} types of CoTs} &   $30$B & $\mathbf{98.8}$  & $\mathbf{93.0}$ &  $\mathbf{95.0}$ \\

        \bottomrule
    \end{tabular}
    }
    \caption{The best performance of individual types of CoTs, and the upper bounds of all types of CoTs (if any of the CoT is correct).}
    \label{tab:ensemble}
\end{table}

\begin{figure}[ht!]
	\begin{subfigure}[b]{0.33\textwidth}
            \adjustbox{max width=0.9\textwidth}{
                \centering
            \includegraphics[width=3.4in]{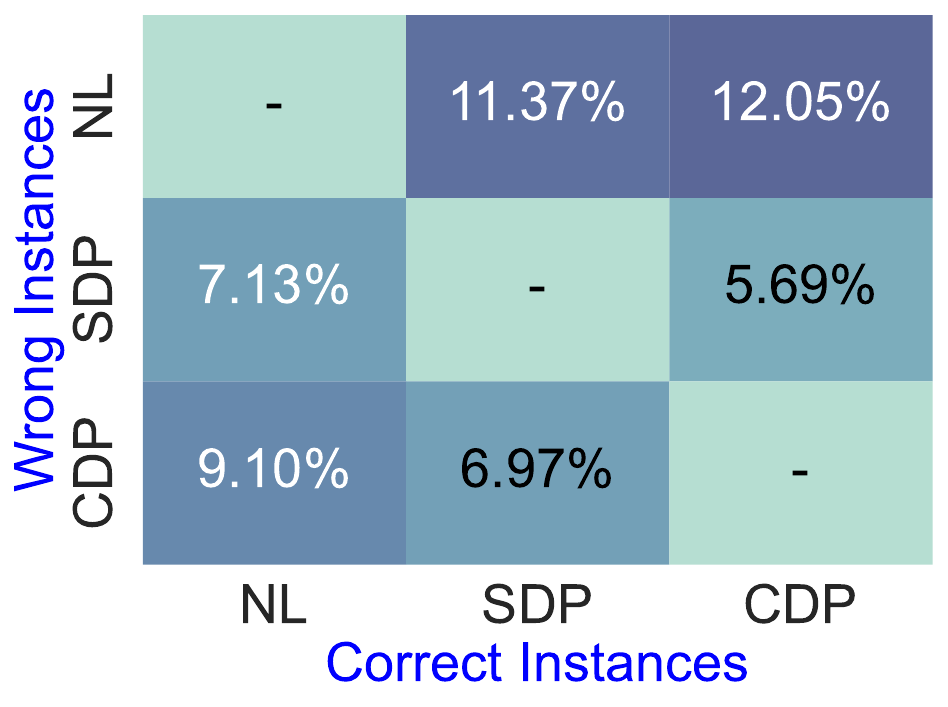}
            }
	\end{subfigure}
         \begin{subfigure}[b]{0.33\textwidth}
            \adjustbox{max width=0.9\textwidth}{
                \centering
            \includegraphics[width=3.4in]{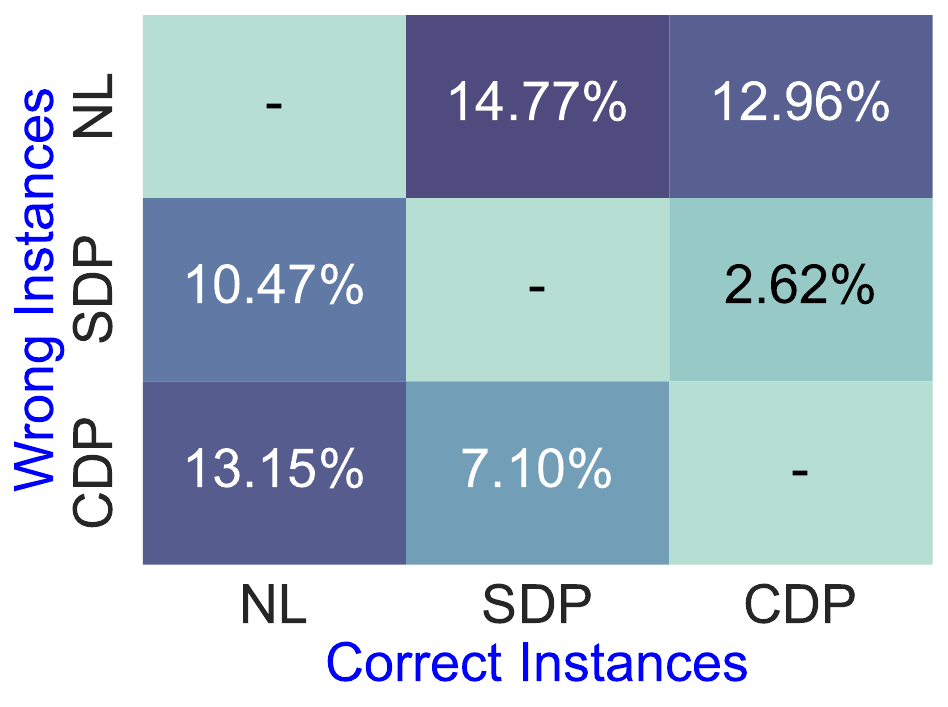}
            }
	\end{subfigure}
        \begin{subfigure}[b]{0.33\textwidth}
            \adjustbox{max width=0.9\textwidth}{
                \centering
            \includegraphics[width=3.4in]{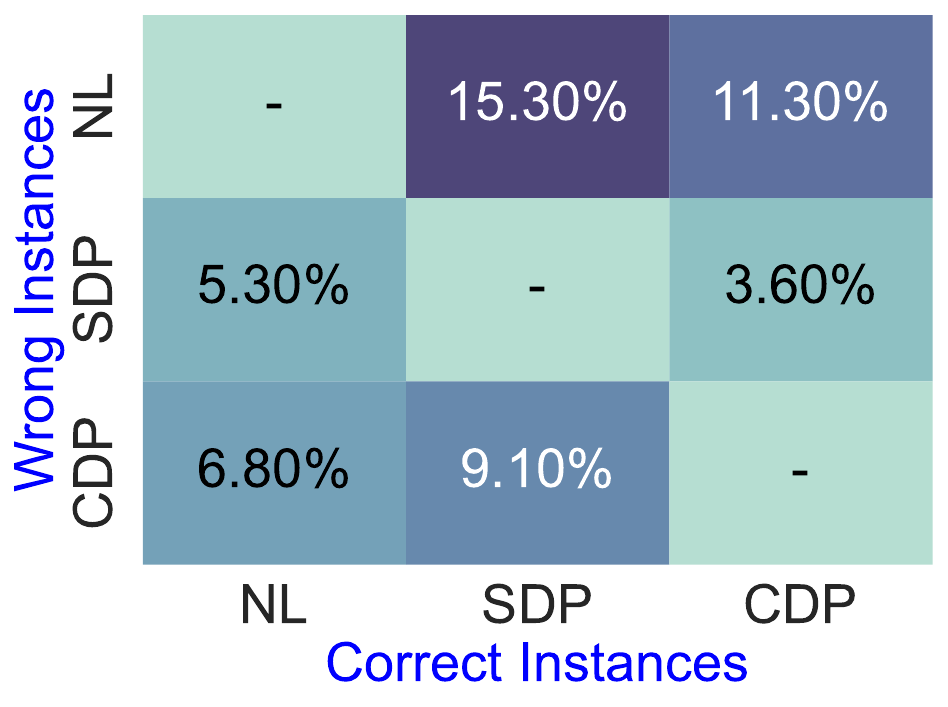}
            }
	\end{subfigure}
	\vspace{-3mm}
	\caption{The percentage of failure cases that are correctly predicted in different CoT types.}
	\label{fig:analysis_compare}
\end{figure}

We also analyze the results of $30$B reward model reranking by comparing different CoT types on the same example (Figure \ref{fig:analysis_compare}). 
Though \textsc{SDP} has overall better performance, a non-negligible amount of failure is correctly solved by \textsc{CDP} or \textsc{NL}.
The same observation applies to the failure cases of \textsc{CDP} or \textsc{NL}, too. 
The above results show that \textsc{CDP}, \textsc{SDP}, and \textsc{NL} have distinct advantages for math-problem-solving. 
We conduct another experiment using a method that treats all three types of CoTs equally during majority voting and reranking. 
For reranking, we train a reward model that is capable of distinguishing and ranking the three types of CoTs. The number of sampled CoT solutions is set to $100$ for fair comparison. 
Specifically, we perform majority voting and reranking on $100$ solutions that contain three types of CoT, in which (\textsc{CDP}, \textsc{SDP} are in Wolfram. 
Table~\ref{tab:unified_result} shows the comparison between the synthetic results and the previous best performance in Table \ref{tab:reranking}. 
The significant improvements suggest that there is still a large potential for a better CoT design that integrates the strengths of all three CoT types.

\begin{table}[ht!]
    \centering
    \adjustbox{max width=0.9\linewidth}{
        \begin{tabular}{llccc}
        \toprule
         \textbf{Size} & \textbf{Method} & \textbf{GSM8K} & \textbf{MathQA} & \textbf{SVAMP}\\ 
        \midrule
        \multirow{4}{*}{$6.7$B} & Majority Voting (Best CoT) & $61.3$ & $72.7$ & $72.9$\\
        & Majority Voting (\textsc{NL} + \textsc{SDP} + \textsc{CDP}) & $67.4$ (\textcolor{darkgrass}{$+6.3$}) & $76.0$ (\textcolor{darkgrass}{$+3.3$}) & $76.0$ (\textcolor{darkgrass}{$+3.1$}) \\
        & Reranking  (Best CoT) & $71.4$ & $73.8$ & $77.3$\\
        & Reranking (\textsc{NL} + \textsc{SDP} + \textsc{CDP}) & $75.4$ (\textcolor{darkgrass}{$+4.0$}) & $79.0$ (\textcolor{darkgrass}{$+5.2$}) & $77.0$ (\textcolor{myred}{$-0.3$})\\
        \midrule
        \multirow{4}{*}{$30$B} & Majority Voting (Best CoT) & $69.8$ & $77.9$ & $79.2$\\
        & Majority Voting (\textsc{NL} + \textsc{SDP} + \textsc{CDP}) & $77.5$ (\textcolor{darkgrass}{$+7.7$})  & $82.6$ (\textcolor{darkgrass}{$+4.7$}) & $81.1$ (\textcolor{darkgrass}{$+1.9$})\\
        & Reranking (Best CoT) & $79.9$& $78.6$&$83.4$\\
        & Reranking (\textsc{NL} + \textsc{SDP} + \textsc{CDP}) & $83.5$ (\textcolor{darkgrass}{$+3.6$})& $83.5$ (\textcolor{darkgrass}{$+4.9$})&$83.9$ (\textcolor{darkgrass}{$+0.5$})\\
        \bottomrule
        \end{tabular}
    }
    \caption{Performance of synthesizing \textsc{CDP}, \textsc{SDP}, and \textsc{NL} CoT types in Wolfram.}
    \label{tab:unified_result}
\end{table}

%% file: 060related.tex
 
Mathematical reasoning through CoT prompting~\citep{wei2022chain}, on large language models~\citep{wei2022emergent}, has experienced significant development in recent years, as evidenced by a large number of CoT methods proposed.
Among them, \cite{uesato2022solving} applied the \textit{process-based} and \textit{outcome-based} reward to score the natural language CoTs on \textsc{GSM8K}\citep{cobbe2021training}, greatly improving problem-solving effectiveness.
\cite{lightman2023lets} enhanced the capability of process-based reward model and achieve significant improvements on the challenging \textsc{MATH} dataset~\citep{hendrycks2021measuring}.
Furthermore, recent research efforts extended simple natural language CoTs, encompassing various approaches designed to enhance and optimize prompting performance. 
Specifically, \cite{fu2022complexity} introduced the concept of \textit{complexity-based prompts}, showing that LLMs favor long reasoning chain, which often leads to superior performance.
Moreover, the methods proposed by \cite{zhou2022least} and \cite{khot2022decomposed} make decomposition of problems into a series of simpler and managable questions.
Similarly, \cite{nye2021show} presented the ``\textit{Scratchpad}'' concept, designed to explicitly present the intermediate calculations to the large language model. 
Although these advancements are significant, ensuring the correctness of CoTs remains a challenge. 

The deterministic nature of programs is increasingly attracting the attention of researchers who use program-aided methods for math problem solving.
\cite{imani2023mathprompter} developed a strategy that ensures answer consistency between programmatic and natural language reasoning, thereby enhancing reliability.
In similar pursuits, both \cite{chen2022program} and \cite{gao2023pal} proposed the use of Python programs as prompts. 
By offloading execution tasks to the Python interpreter, they were able to mitigate issues related to incorrect reasoning or calculations. 
The programs employed in these approaches are similar to our self-describing programs, where variables are represented using natural language. 
\cite{zhou2023solving} further combined the natural language and program by making use of the code interpreter in GPT-4~\citep{openai2023gpt4}. 
Concurrently, research by \cite{drori2022neural} and \cite{li2022competition} demonstrated the effectiveness of generating purely symbolic Python programs to address MATH questions~\citep{hendrycks2021measuring} in programming competitions.
\cite{he2023solving} enabled declarative reasoning in a program by embedding symbolic expressions into natural language prompts.
In response to the diversity of program CoT types, our work aims to provide a comprehensive analysis and comparison of the representations. 
Our objective is to uncover their distinctive characteristics and potential advantages.

%% file: 070conclusion.tex

We have conducted a comprehensive study of chain-of-thought design for math problem solving, including natural language and program CoTs. 
We categorize the program CoTs into non-describing program, self-describing program, and comment-describing program. 
Through extensive experiments on \textsc{GSM8K}, \textsc{MathQA} and \textsc{SVAMP}, we find that the self-describing program often achieves the best performance and outperforms the few-shot prompting by {GPT-3.5-turbo}. 
It is better to use program CoTs than natural language CoTs for math problem solving.
Self-describing program and comment-describing program perform better than non-describing program.
Among the first two, self-describing program works better than comment-describing program. 
The program CoTs in Python work better than the program CoTs in Wolfram.
We hope our experimental findings will provide valuable insights for the future design of chain-of-thought in math problem solving.

%% file: 100appendix.tex

\begin{table*}[ht!]
\begin{center}
\begin{tabular}{lccc}
\toprule
\bf Hyperparam  & \bf Supervised Fine-tuning & \bf Reward Modeling \\
\bf   & \bf ($6B/30B$) & \bf ($6B/30B$) \\
\midrule 
Framework & Megatron-Deepspeed & Pytorch \\
GPUs & 16 A100 & 8 A100 \\
Maximum sequence length & $1024$ & $700$ \\
Learning rate & $1e^{-5}/2e^{-6}$ & $1e^{-6}$ \\
Batch size & $48$ & $48$ \\
Warmup Steps & $10\%$ & $10\%$ \\
Weight Decay & $0.01$ & $0.01$ \\
Epoch & $40$ & $3$\\
Learning Rate Decay & Linear & Linear \\
Adam $\epsilon$ & $1e-6$ & $1e-6$ \\
Adam $\beta_1$ & $0.9$ & $0.9$ \\
Adam $\beta_2$ & $0.98$ & $0.98$ \\
Gradient Clipping & $1.0$ & $1.0$ \\
\bottomrule
\end{tabular}
\end{center}
\caption{
Hyperparameters for Supervised Fine-Tuning and Reward Modeling for 6B and 30B parameters model scale.
}
\label{tab:SFT and RM hyperparameter}
\end{table*}